\documentclass{article}

\usepackage[preprint]{neurips_2025}

\usepackage[utf8]{inputenc} 
\usepackage[T1]{fontenc}    
\usepackage{hyperref}       
\usepackage{url}            
\usepackage{booktabs}       
\usepackage{amsfonts}       
\usepackage{nicefrac}       
\usepackage{microtype}      
\usepackage{inconsolata}
\usepackage{caption}
\usepackage{amsmath}
\usepackage{amssymb}
\usepackage{mathtools}
\usepackage{enumitem}
\usepackage{makecell} 
\usepackage[usestackEOL]{stackengine}
\usepackage{graphicx}
\usepackage{capt-of}
\usepackage{booktabs}
\usepackage{varwidth}
\usepackage[table,dvipsnames,svgnames]{xcolor}

\usepackage{tikz}
\usetikzlibrary{shadows}
\usepackage{float}
\usepackage{caption}
\usepackage{subcaption}
\usepackage{xspace}
\usepackage{svg}

\usepackage{import}
\usepackage{animate}
\usepackage{arydshln}
\usepackage{multirow}
\usepackage[normalem]{ulem}
\usepackage[most]{tcolorbox}
\usepackage{colortbl}
\usepackage{tcolorbox}
\usepackage{pifont}
\usepackage{alltt}
\definecolor{titlegray}{rgb}{0.4, 0.4, 0.4} 
\definecolor{contentgray}{rgb}{0.95, 0.95, 0.95} 
\usepackage{algorithm}
\usepackage{algpseudocode}

\newcommand{\model}{\text{C3}\xspace}

\definecolor{carolinablue}{rgb}{0.6, 0.73, 0.89}
\definecolor{mildgreen}{rgb}{0.85, 0.98, 0.80}
\definecolor{beautycolor}{rgb}{0.91, 0.75, 0.96} 
\definecolor{fallacycolor}{rgb}{0.85, 0.95, 1}
\definecolor{gendercolor}{rgb}{1, 0.85, 0.85}
\definecolor{brightyellow}{RGB}{255, 255, 100}
\definecolor{boxcolor}{RGB}{51,51,153}
\definecolor{lightgreen}{rgb}{0.56, 0.93, 0.56}
\definecolor{citeblue}{HTML}{0064E0}

\hypersetup{
    colorlinks=true,
    citecolor=citeblue,
    linkcolor=red, 
    urlcolor=citeblue 
}

\definecolor{deepblue}{RGB}{0, 0, 139}

\title{Context Cascade Compression: Exploring the Upper Limits of Text Compression}

\newtcolorbox{questionbanner}{
  colback=blue!10!white,    
  colframe=blue!80!black,   
  width=\textwidth,
  arc=4mm,                  
  boxrule=1pt,              
  fonttitle=\bfseries,
  title=Question,
}
\newtcolorbox{promptbox}[1]{
  colback=contentgray,      
  colframe=titlegray,       
  colbacktitle=titlegray,   
  coltitle=white,           
  title={#1}, 
  arc=4mm,                  
  rounded corners=northwest, 
  rounded corners=northeast, 
  sharp corners=south,      
  boxrule=1pt,              
  fonttitle=\bfseries,      
}

\definecolor{questionbg}{RGB}{240, 248, 255}  
\definecolor{answerbg}{RGB}{245, 255, 250}   
\definecolor{bordercolor}{RGB}{100, 149, 237} 
\definecolor{titlecolor}{RGB}{25, 25, 112}    

\newtcolorbox{vqaexample}[2][]{
    enhanced,
    breakable,
    colback=white,
    colframe=bordercolor,
    boxrule=1.5pt,
    arc=4pt,
    outer arc=4pt,
    left=8pt,
    right=8pt,
    top=8pt,
    bottom=8pt,
    drop shadow={shadow xshift=0.5mm, shadow yshift=-0.5mm, opacity=0.3},
    overlay={
        \node[
            anchor=north east,
            xshift=-3pt,
            yshift=-3pt,
            fill=bordercolor!80,
            text=white,
            font=\bfseries,
            rounded corners=2pt,
            inner sep=4pt,
            minimum height=1.2em,
            align=center
        ] at (frame.north east) {#2};
    },
    #1
}

\author{
    Fanfan Liu,\hspace{0.3em}
    Haibo Qiu\hspace{0.3em}
    \\
    \\
}

\begin{document}

\maketitle
\vspace{-5mm}

\begin{abstract}
Million-level token inputs in long-context tasks pose significant computational and memory challenges for Large Language Models (LLMs). Recently, DeepSeek-OCR conducted research into the feasibility of Contexts Optical Compression and achieved preliminary results. Inspired by this, we introduce \textbf{Context Cascade Compression (\model)} to explore the upper limits of text compression. Our method cascades two LLMs of different sizes to handle the compression and decoding tasks. Specifically, a small LLM, acting as the first stage, performs text compression by condensing a long context into a set of latent tokens (e.g., 32 or 64 in length), achieving a high ratio of text tokens to latent tokens. A large LLM, as the second stage, then executes the decoding task on this compressed context. Experiments show that at a 20x compression ratio (where the number of text tokens is 20 times the number of latent tokens), our model achieves 98\% decoding accuracy, compared to approximately 60\% for DeepSeek-OCR. When we further increase the compression ratio to 40x, the accuracy is maintained at around 93\%. This indicates that in the domain of context compression, \model Compression demonstrates superior performance and feasibility over optical character compression. C3 uses a simpler, pure-text pipeline that ignores factors like layout, color, and information loss from a visual encoder. This also suggests a potential upper bound for compression ratios in future work on optical character compression, OCR, and related fields. Codes and model weights are publicly accessible at \href{https://github.com/liufanfanlff/C3-Context-Cascade-Compression}{https://github.com/liufanfanlff/C3-Context-Cascade-Compression} 
\end{abstract}

\begin{figure}[h]
    \centering
    \includegraphics[width=\linewidth]{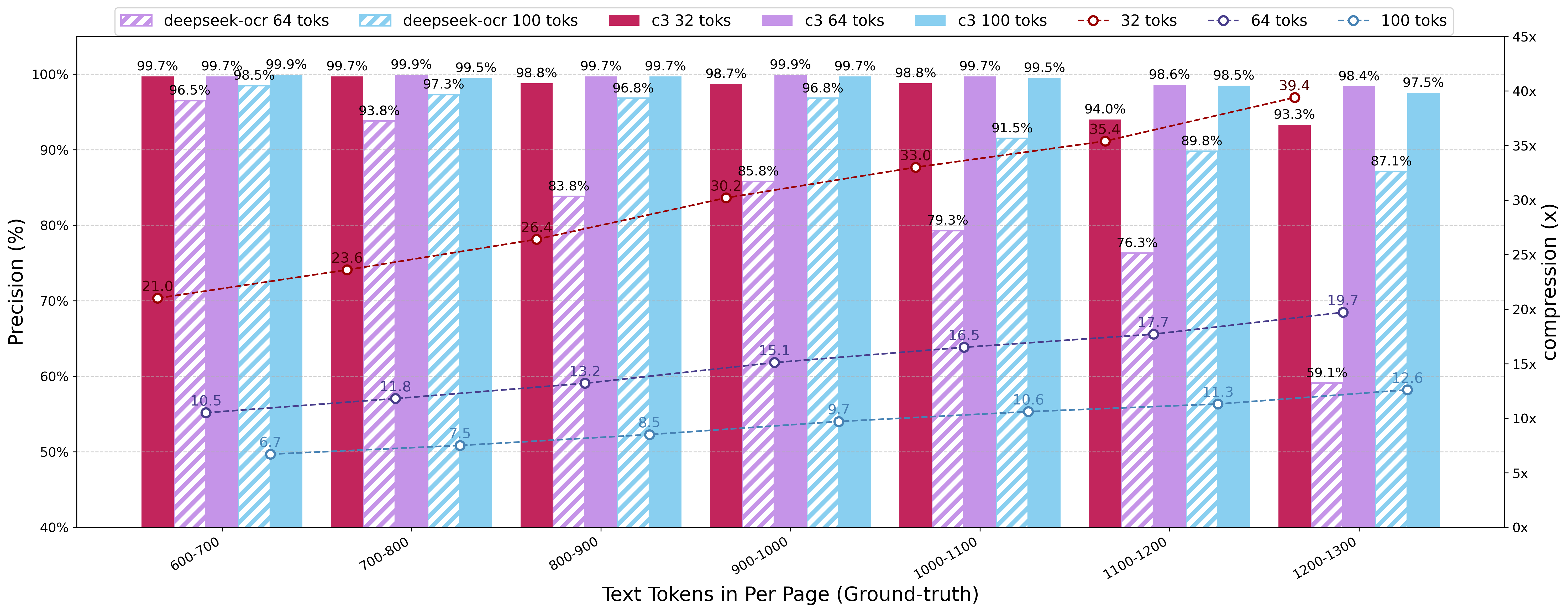} 
    \caption{Reconstruction precision and compression ratio of C3 versus Deepseek-OCR on the Fox benchmark. Bars represent precision (left axis), while lines indicate the compression ratio (right axis). C3 (solid bars) demonstrates significantly higher precision than Deepseek-OCR (striped bars) across all tested latent token counts (32, 64, and 100).}
    \label{fig:c3vsdpskocr}
\end{figure}

\newpage
\begin{figure}[h]
    \centering
    \includegraphics[width=\linewidth]{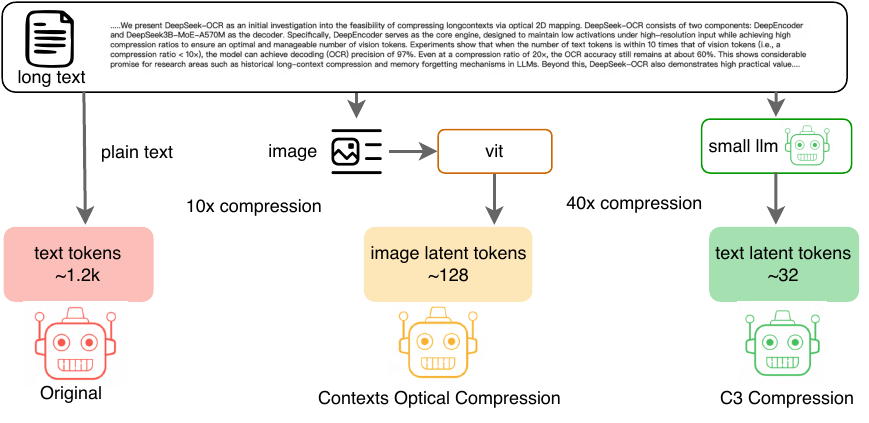} 
    \caption{Conceptual comparison of context compression pipelines. The figure contrasts three methods: (1) Original: The baseline tokenization of plain text, leading to a large token count (\textasciitilde1.2k). (2) Contexts Optical Compression: An indirect method where text is converted to an image and encoded by a ViT, achieving \textasciitilde10x compression into \textasciitilde128 latent tokens. (3) C3 Compression: Our proposed direct method, where a small LLM compresses the text into a minimal set of \textasciitilde32 latent tokens, achieving \textasciitilde40x compression.}
    \label{fig:introduction}
\end{figure}
\section{Introduction}
\label{section:introduction}
Recently, Large Language Models (LLMs) ~\citep{GPT4} have achieved significant advancements in various fields. As the capabilities of LLMs continue to grow and their application scenarios expand, the computational and memory challenges they face in long-context scenarios have become increasingly evident. Recent research has explored several effective methods to alleviate this issue. An approach is to modify the attention mechanism, for example, using sparse~\citep{deepseekv32exp2025} or linear attention~\citep{qwen3next2025}, to reduce the quadratic complexity of self-attention and improve computational efficiency. However, as the context length grows to hundreds of thousands of tokens, the overall overhead remains substantial because the number of tokens is unchanged. Alternatively, retrieval-augmented methods~\citep{lewis2021retrievalaugmentedgenerationknowledgeintensivenlp} shorten the input length through external retrieval, but this is essentially a form of lossy information compression and introduces additional latency. Another novel approach uses visual modality as an efficient medium for compressing textual information~\citep{wei2025deepseek}; an image containing a document's text can represent the information with far fewer tokens than the text itself. Based on this insight, DeepSeek-OCR~\citep{wei2025deepseek} employs Optical Character Recognition (OCR) as an intermediate modality to bridge vision and language. It deeply explores this compression paradigm, achieving a 10x text token compression rate while maintaining a 97\% decoding accuracy.

We revisit the entire pipeline of DeepSeek-OCR and find that when using the visual modality as a compression medium for text, as shown in figure  \ref{fig:introduction}, the LLM process shifts from $Text \rightarrow Text Tokens \rightarrow LLM$ to $Text \rightarrow Image \rightarrow Visual Latent Tokens \rightarrow LLM$. In this new pipeline, we hypothesize that information compression may not stem primarily from $Text \rightarrow Image$ conversion, but rather from the fact that latent tokens are a more efficient information representation compared to discrete text tokens. Based on this hypothesis, we propose a novel context compression paradigm: Context Cascade Compression (\model). In \model, we directly cascade two LLMs of different sizes. The first, smaller LLM compresses discrete text tokens into textual latent tokens, functioning similarly to a visual encoder in a VLM. The second, larger LLM then performs the downstream task. (For this paper, we focus on information preservation, tasking the model to reconstruct the original text from the compressed latent tokens, a process equivalent to the OCR task of recognizing text from visual latent tokens). Compared to DeepSeek-OCR, \model employs a more streamlined pipeline for pure text compression, as it disregards other factors in optical character compression, such as layout, color, and information loss from the visual encoder. Consequently, \model achieves a compression rate far exceeding that of optical methods while maintaining the same decoding accuracy. It also provides a potential theoretical upper bound for the ratio of text tokens to visual tokens in future work on optical character compression and OCR.

In summary, our main contributions are as follows:
\begin{itemize}[leftmargin=0.8cm]
 \item We propose Context Cascade Compression (C3), a novel architecture to achieve highly efficient compression of long contexts.
 
 \item We achieve a performance that far surpasses optical character compression. At a 40x compression ratio (where the number of text tokens is 40 times the number of latent tokens), our model maintains a decoding accuracy of 93\%, compared to approximately 10x for DeepSeek-OCR. Our architecture offers a more efficient and feasible approach for future ultra-long context inputs.
 \item We analyze the forgetting pattern of C3 and find that it exhibits sequential information loss, with errors tending to concentrate at the end of the text. This process is highly analogous to the decay of human memory.

\end{itemize}
In conclusion, our work provides a preliminary exploration into the upper limits of text compression rates and demonstrates that direct text compression is a more efficient method than optical character compression. Moreover, our method presents a viable solution for handling future ultra-long context inputs. From another perspective, our architecture can also serve as an autoencoder structure for diffusion language models and latent auto-regressive models, enabling the conversion of variable-length contexts into fixed-length latent tokens.
\section{Related Work}
\label{section:related}
In recent years, Large Language Models (LLMs) have demonstrated remarkable capabilities. However, their core component, the self-attention mechanism, exhibits quadratic computational and memory complexity with respect to the input sequence length. This has severely limited the context length that models can effectively handle. To overcome this bottleneck, both academia and industry have explored various techniques for long-context processing and compression. Our work is primarily related to context compression, with a specific focus on the emerging field of contexts optical compression.
\paragraph{Contexts Compression.}
Handling long contexts is a frontier in current LLM research. Existing methods can be broadly categorized as follows:

Efficient Attention Mechanisms: A significant body of research focuses on optimizing the attention mechanism at an architectural level to reduce its quadratic complexity. For instance, sparse attention methods, such as Longformer~\citep{Beltagy2020Longformer} and BigBird~\citep{zaheer2020bigbird}, reduce computational load by limiting the attention scope of each token. Other approaches, like Linear Attention~\citep{katharopoulos2020transformers} and State-Space Models (e.g., Mamba~\citep{mamba}), attempt to approximate the performance of standard attention with linear or near-linear complexity. Although effective, these methods often require fundamental architectural modifications and may trade off some performance on certain tasks.

Retrieval-Augmented Generation: RAG~\citep{lewis2021retrievalaugmentedgenerationknowledgeintensivenlp} adopts a different strategy by storing long contexts in an external knowledge base (e.g., a vector database) and retrieving only the most relevant snippets to inject into the model's input at inference time. This approach excels in information-retrieval tasks but is inherently a ``loss'' compression method, as it relies on the accuracy of the retrieval step and may lose global context or inter-snippet relationships.

Pluggable Module:
Drawing parallels to modules like Q-Former in VLMs, some prior work has also explored prepending a carefully designed external module to reduce the length of input tokens for the main LLM. Notable examples include~\citep{ge2023context,wang2024context}. However, these modules have generally demonstrated severely limited compression performance.

Memory and Summarization: Other works have attempted to compress context by generating intermediate summaries or memory tokens. For example, some models periodically generate summaries of long texts to represent historical information ~\citep{bulatov2022recurrent}. Our work, C3, shares a philosophy similar with these methods in that it generates a compact representation of the context. However, C3 is distinct in its approach: rather than generating a human-readable text summary, we train a dedicated encoder LLM to directly distill textual information into a fixed-length set of non-interpretable latent tokens. This end-to-end, latent-space compression paradigm potentially preserves richer semantic information and provides a more direct and efficient input for the downstream decoder compared to generative summarization.

\paragraph{Contexts Optical Compression.}

Using the visual modality as a compression medium for text is a novel and promising research direction, which we term "Contexts Optical Compression."~\citep{wei2025deepseek}

The core idea is to render long text into one or more images and then leverage a powerful visual encoder to compress the high-dimensional pixel information into a series of visual tokens. The theoretical basis for this approach is that visual encoders have demonstrated exceptional ability in extracting dense features from complex images, and a text-rendered image can be viewed as a highly structured visual signal.

The most recent representative work in this area are DeepSeek-OCR~\citep{wei2025deepseek}, Glyph~\citep{cheng2025glyphscalingcontextwindows}. This model employs the OCR task as a bridge between vision and language, compressing long text by rendering it into images. It achieves a token compression rate of up to 10x while maintaining high decoding accuracy, proving the feasibility of the optical compression pathway. Our work, C3, offers new insights and improvements upon this paradigm.

\begin{figure}[t]
    \centering
    \includegraphics[width=\linewidth]{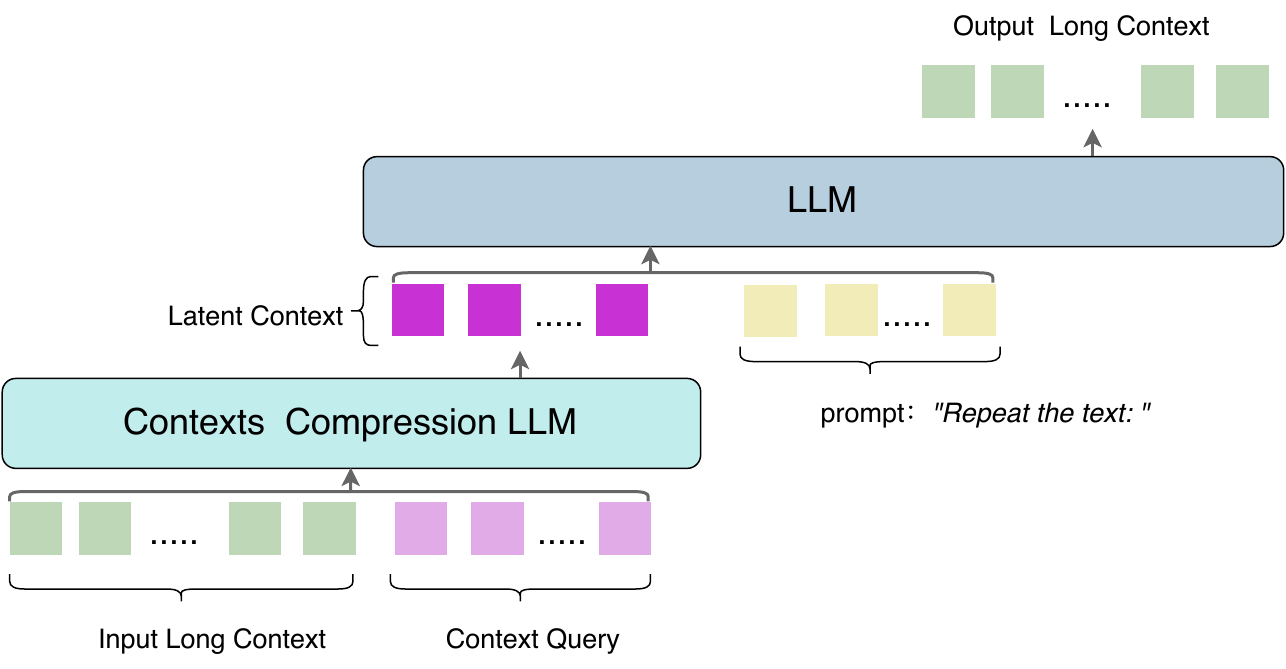} 
    \caption{An overview of the C3, which utilizes a cascaded two-LLM design. A smaller encoder LLM compresses a variable-length Input Long Context into a fixed-length Latent Context guided by learnable Context Query tokens. Subsequently, a larger decoder LLM uses this compact Latent Context and a prompt to perform the downstream task, such as reconstructing the original text. }
    \label{fig:method}
\end{figure}

\section{Method}
\label{section:method}

\subsection{Architecture}
As shown in Figure \ref{fig:method}, C3 employs a cascaded two-LLM architecture, comprising a context compression encoder LLM and a decoder LLM. The context compression encoder is responsible for information compression, transforming text tokens into latent tokens. The decoder then utilizes these latent tokens and a given prompt to generate the desired output. Specifically, we use Qwen2.5 1.5B~\citep{qwen2025qwen25technicalreport} as the context compression encoder and Qwen2.5 3B~\citep{qwen2025qwen25technicalreport}  as the decoder. In the following sections, we will delve into the details of the model components, data engineering, and training methodology.

\subsection{Context Compression Encoder LLM}

In this section, we detail the architecture of our text compression encoder. The primary objective is to formulate a component that is both architecturally concise and computationally efficient for compressing long textual sequences. We posit that modern, large-scale, pre-trained language models, by virtue of their extensive training, have already developed a sophisticated capacity for information extraction, semantic understanding, and summarization. Capitalizing on this inherent capability, we eschew designing a compression module from scratch and instead directly adapt a pre-trained LLM for this purpose.

Specifically, the architectural backbone of our context compression encoder is initialized using the weights of a pre-trained Qwen2.5 1.5B model. To facilitate the compression, we introduce a set of learnable embeddings designated as the context query.  This query is materialized as a trainable tensor Q of dimensions $N \times D$, where $N$ represents the fixed number of tokens desired for the output latent context, and $D$ corresponds to the hidden dimension of the Qwen2.5 1.5B model's embedding layer.

The input sequence fed to the encoder is a concatenation of the original long context (as text tokens) and context query embeddings.
Crucially, the model processes this hybrid sequence uniformly. The context query embeddings are treated identically to standard text tokens within the model's self-attention mechanism. No architectural modifications, such as introducing cross-attention layers, are required. The entire forward pass relies solely on the model's native causal attention mechanism. This design choice ensures simplicity and leverages the full expressive power of the pre-trained Transformer architecture.Upon completion of the forward pass, the final layer's output hidden states corresponding to the positions of the context query tokens are extracted. This resulting tensor, of shape $N \times D$, serves as an efficient and dense representation of the original text. This output constitutes the latent context, which is then passed to the downstream decoder.

\subsection{Decoder LLM}

Following the established paradigm in advanced generative models, such as Vision-Language Models (VLMs) and recent OCR systems~\citep{wei2024general}, we employ a Large Language Model (LLM) as the decoder component. The primary function of this decoder is to interpret the dense, compressed latent context provided by the encoder and to generate a coherent textual output that fulfills a specified downstream task.

For the scope of this paper, our investigation is focused exclusively on the fundamental task of text reconstruction. This task serves as a direct and rigorous benchmark for evaluating the information fidelity of our C3 compression architecture. By tasking the model with perfectly recreating the original input text from its compressed representation, we can quantitatively measure the degree of information preserved throughout the compression-decompression cycle. This setup provides a clear, objective measure of the compression's ``lossiness'' and establishes a baseline for the model's capabilities.

Operationally, the input to the decoder is a concatenated sequence comprising the latent context and a task-specific prompt. When performing the reconstruction task, we use the explicit instruction \textit{"repeat the text: "}. The decoder is then trained to auto-regressively generate a token sequence that is identical to the original, ground-truth text, thereby demonstrating that the semantic integrity of the input has been successfully maintained within the latent context.

\section{Experiment}
\label{section:experiment}

\subsection{Data}
We trained our model using ground truth from OCR data. We collected 1 million pages of diverse OCR data from the internet, primarily in English and Chinese. The dataset is predominantly composed of document-related data. Notably, our approach demonstrates considerable robustness with respect to data preparation. We found that extensive data engineering or complex curation was not necessary. A straightforward aggregation of text samples of diverse lengths proved sufficient to train the model effectively, indicating that our proposed architecture is inherently resilient to variations in input length and does not depend on meticulous data preprocessing to achieve strong performance.

\subsection{Training Setup}

The model was trained on a high-performance computing cluster equipped with 8 NVIDIA H800 GPUs. We configured the training with a per-device batch size of 2 and employed 16 gradient accumulation steps. This configuration resulted in an effective global batch size of 256 (2 per device × 8 GPUs × 16 accumulation steps). The model was optimized using AdamW~\citep{loshchilov2019decoupledweightdecayregularization} optimizer with a peak learning rate of 1e-5. A cosine~\citep{loshchilov2017sgdrstochasticgradientdescent} learning rate scheduler was employed to anneal the learning rate over the course of training, preceded by a linear warmup~\citep{vaswani2017attention} phase over the initial 100 steps. The entire training process was conducted for a total of 40,000 steps.


\subsection{Context Compression Study}



\subsubsection{Quantitative}
To rigorously evaluate the compression and reconstruction fidelity of our C3 architecture, we follow the experimental protocol established by DeepSeek-OCR , thereby enabling a direct and fair comparison of performance. We adopt the Fox~\citep{liu2024focus} benchmark, to assess our model's capabilities.
Specifically, our evaluation on the English document portion of the Fox benchmark. The ground-truth texts are tokenized using the official Qwen tokenizer, and for our test set, we select documents with token counts ranging from 600 to 1300 tokens. This selection ensures a consistent basis for comparing the performance of our text-to-latent paradigm against the optical compression paradigm. While minor discrepancies in token counts may arise due to differences between our tokenizer and the one used in the original work, we posit that this effect on the final results is negligible.
In our experiments, we benchmark performance at multiple compression levels. To align with the baseline, we evaluate settings where the contexts are compressed into 64 and 100 latent tokens. Furthermore, to explore the upper limits of our compression method, we introduce a more aggressive setting, evaluating the performance with a highly compressed representation of only 32 latent tokens. Across all evaluations, the text reconstruction task is initiated using the prompt \textit{"repeat the text: "}.

The experimental results, presented in Table \ref{tab:c3_vs_deepseek_centered}, provide a comprehensive comparison between our C3 architecture and the Deepseek-OCR baseline on the Fox benchmark. The data unequivocally demonstrates that C3's direct text-to-latent compression paradigm significantly outperforms the optical compression approach across all tested conditions, establishing a new state-of-the-art in high-fidelity context compression.
A detailed analysis of the results reveals several key insights. In the Vision Tokens = 100 setting, where Deepseek-OCR achieves its peak performance of approximately 97\% precision at a ~7-10x compression ratio, our C3 model consistently maintains a precision of over 99.5\% under similar compression rates. This indicates that even at moderate compression levels, our direct approach offers superior information preservation.
The performance gap widens dramatically at higher compression ratios, particularly in the more aggressive Vision Tokens = 64 setting. While Deepseek-OCR's performance begins to degrade notably as the compression ratio exceeds 10x (dropping to 83.8\% at 13.2x), C3 sustains near-perfect reconstruction. Most strikingly, when compressing the context by nearly 20x (1200-1300 text tokens into 64 latent tokens), C3 maintains an exceptional precision of 98.4\%. In stark contrast, the performance of Deepseek-OCR under similar conditions plummets to 59.1\%. This result provides compelling evidence that our method is far more robust and effective for extreme context compression.

The superior performance of C3 can be attributed to its fundamental architectural design. The Deepseek-OCR analysis hypothesizes that its performance decline is due to factors like ``complex layout'' and ``image blurring at lower resolutions''—inherent limitations of the optical pathway. Our C3 paradigm, by operating directly in the textual domain, is entirely immune to these visual-domain artifacts. It avoids the information loss associated with rendering text to pixels and then encoding those pixels. Instead, it leverages a pre-trained LLM's powerful semantic understanding to distill textual information directly into an efficient latent representation.

\begin{table}[htbp]
\centering
\caption{Performance comparison between Deepseek-OCR (DS-OCR) and C3 at different compression levels. The table reports reconstruction precision and compression ratios for documents of varying lengths, using 64 and 100 latent tokens.}
\label{tab:c3_vs_deepseek_centered}
\begin{tabular}{ccccccc} 
\toprule
 & \multicolumn{3}{c}{\textbf{Latent Tokens = 64}} & \multicolumn{3}{c}{\textbf{Latent Tokens = 100}} \\
\cmidrule(lr){2-4} \cmidrule(lr){5-7}
\textbf{Text Tokens} & \textbf{C3} & \textbf{DS-OCR} & \textbf{Compression} & \textbf{C3} & \textbf{DS-OCR} & \textbf{Compression} \\
\midrule
600-700   & 99.7\% & 96.5\% & 10.5$\times$ & 99.9\% & 98.5\% & 6.7$\times$ \\
700-800   & 99.9\% & 93.8\% & 11.8$\times$ & 99.5\% & 97.3\% & 7.5$\times$ \\
800-900   & 99.7\% & 83.8\% & 13.2$\times$ & 99.7\% & 96.8\% & 8.5$\times$ \\
900-1000  & 99.9\% & 85.9\% & 15.1$\times$ & 99.7\% & 96.8\% & 9.7$\times$ \\
1000-1100 & 99.7\% & 79.3\% & 16.5$\times$ & 99.5\% & 91.5\% & 10.6$\times$ \\
1100-1200 & 98.6\% & 76.4\% & 17.7$\times$ & 98.5\% & 89.8\% & 11.3$\times$ \\
1200-1300 & 98.4\% & 59.1\% & 19.7$\times$ & 97.5\% & 87.1\% & 12.6$\times$ \\
\bottomrule
\end{tabular}
\end{table}

To further probe the upper limits of our C3 architecture, we conducted a more aggressive set of experiments, compressing long contexts into just 32 latent tokens. The results, detailed in Table \ref{tab:32-Latent-Tokens}, are highly compelling and underscore the remarkable efficiency and robustness of our direct text-to-latent paradigm.The data reveals that even under these extreme compression settings, C3 maintains an exceptionally high level of reconstruction fidelity. 
Even when compressing documents by over 30x (for texts of 900-1000 tokens), C3 sustains a precision of 98.8\%. At the far end of our test, compressing 1200-1300 tokens by nearly 40x, the model still retains a remarkable 93.3\% precision.
Deepseek-OCR, which, as shown previously (in Table \ref{tab:c3_vs_deepseek_centered}), degrade to approximately 60\% precision at a mere 20x compression ratio. Our C3 model not only surpasses that benchmark but does so while operating at nearly double the compression rate.

These findings conclusively demonstrate the advantage of the C3 architecture. By avoiding the information bottlenecks and potential artifacts inherent in the visual modality (e.g., image resolution limits, layout complexity), our method gracefully handles extreme compression with minimal information loss. The results from this aggressive test case solidify our claim that direct text-to-latent compression is a fundamentally more efficient and powerful paradigm than its optical counterparts.

\begin{table}[htbp]
\centering
\caption{Precision and Compression Ratio at 32 Latent Tokens}
\label{tab:32-Latent-Tokens}
\begin{tabular}{ccc} 
\toprule
\multicolumn{3}{c}{\textbf{Latent Tokens = 32}} \\
\cmidrule(lr){1-3} 
\textbf{Text Tokens} & \textbf{precision}  & \textbf{Compression}  \\
\midrule
600-700   & 99.7\%  & 21.0$\times$  \\
700-800   & 99.7\%  & 23.6$\times$  \\
800-900   & 98.8\%  & 26.4$\times$  \\
900-1000  & 98.7\%  & 30.2$\times$  \\
1000-1100 & 98.8\%  & 31.0$\times$  \\
1100-1200 & 94.0\%  & 35.4$\times$  \\
1200-1300 & 93.3\%  & 39.4$\times$  \\
\bottomrule
\end{tabular}
\end{table}

\subsubsection{Qualitative}
\begin{figure}[ht]
    \centering
    \includegraphics[width=\linewidth]{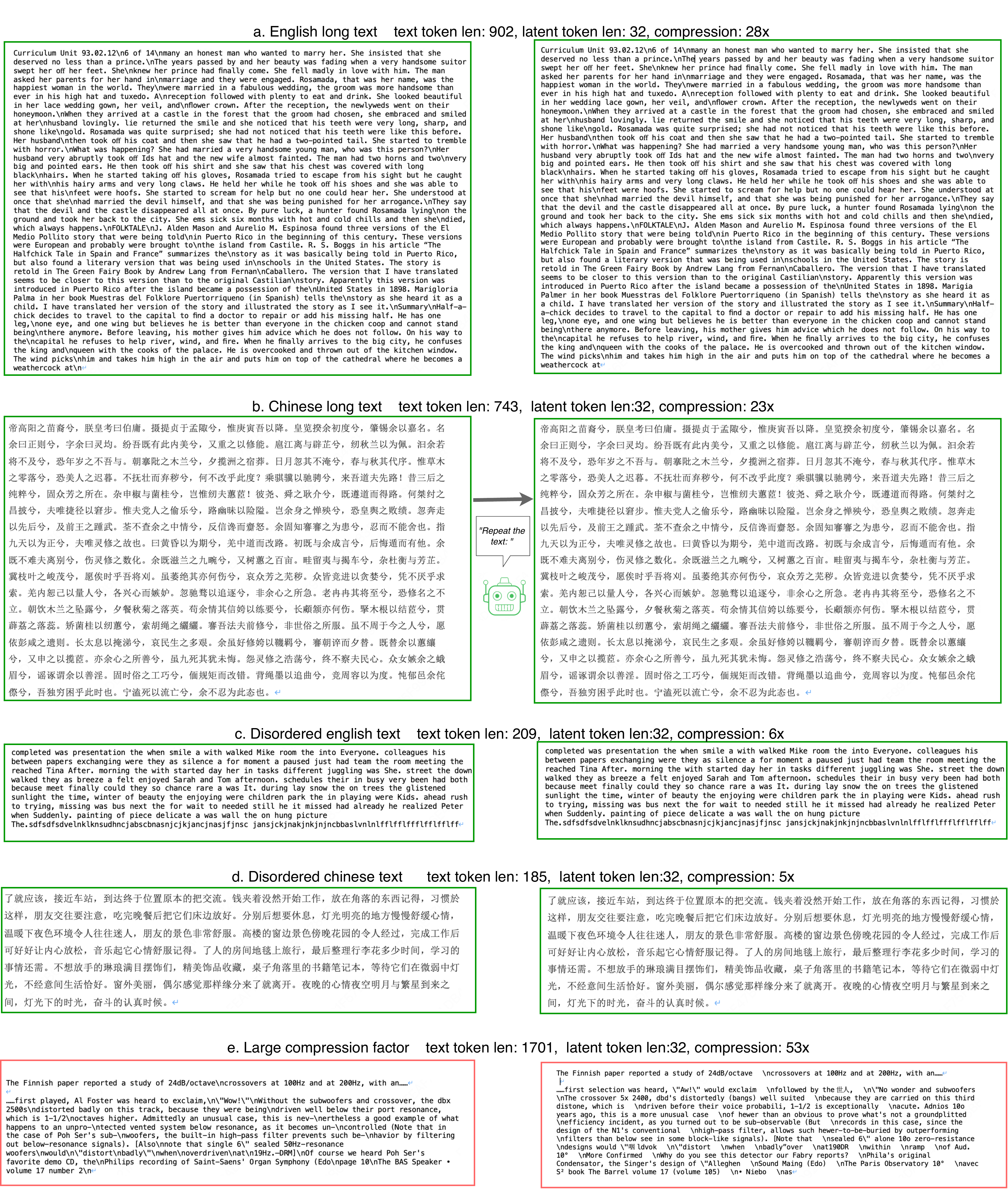} 
    \caption{Qualitative results of text reconstruction using C3 at an extreme compression level (32 latent tokens). Each panel displays the original long text on the left and the model's reconstructed output on the right. The examples showcase the model's high-fidelity performance across diverse scenarios, including: (a) standard English prose, (b) classical Chinese, (c) English text containing non-semantic random characters, and (d) structurally disordered Chinese text. The near-perfect reconstruction in all cases highlights C3's capability for near-lossless compression. Furthermore, we present an analysis of failure cases that occur under extreme compression ratios. (e) A key observation is that in these ``bad cases'', reconstruction errors tend to be concentrated in the latter half of the original text.}
    \label{fig:example}
\end{figure}

To provide a more intuitive understanding of our model's capabilities, we present several qualitative case studies in Figure \ref{fig:example}. These examples showcase C3's reconstruction performance at an extreme compression level, where long contexts are compressed into just 32 latent tokens.

The first two examples (top row) demonstrate the model's high-fidelity reconstruction on well-structured English prose and classical Chinese. In both instances, the model achieves a near-perfect, verbatim copy of the original input, confirming its efficacy on standard, coherent text.

More notably, the bottom two examples test the model's robustness against unconventional inputs: an English paragraph interspersed with non-semantic random characters, and a structurally disordered Chinese paragraph. In both challenging scenarios, C3 successfully reconstructs the input verbatim, including the random characters and the disordered sentence structure. It effectively learns to ``zip'' and ``unzip'' the textual data, regardless of its semantic coherence.

\begin{figure}[t]
    \centering
    \includegraphics[width=\linewidth]{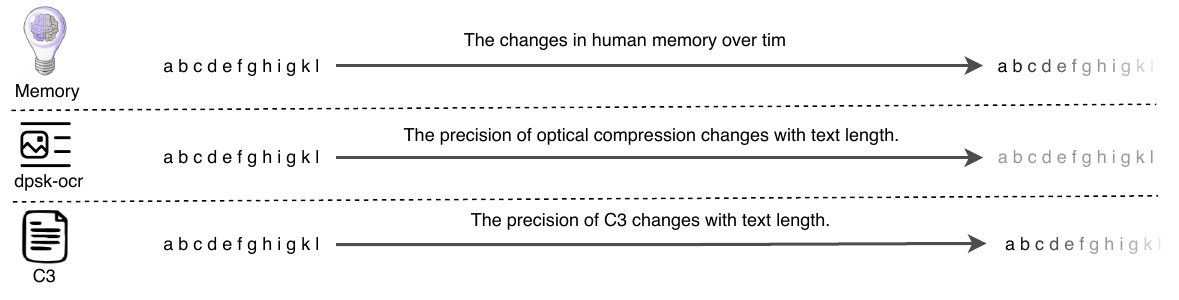} 
    \caption{An analogy of information loss patterns. This figure contrasts two failure modes. Optical compression (middle) leads to a uniform 'blurring' of the entire context. In contrast, C3's information loss is sequential (bottom), fading from the end, which is analogous to the process of human memory decay (top).}
    \label{fig:Qualitative}
\end{figure}

In addition to these successful cases, it is instructive to examine the model's failure modes under extreme pressure. When errors do occur, they tend to be concentrated in the latter of the text. 
In contrast to optical compression methods, which exhibit a uniform degradation of information across the entire text as the number of latent tokens decreases, our approach demonstrates a sequential information loss. Fidelity is often perfectly preserved for the initial parts of the text, while errors gradually manifest towards the end. This behavior is more analogous to the human process of forgetting, where information loss occurs progressively over time rather than as a diffuse, simultaneous blur. As show in Figure \ref{fig:Qualitative}, this suggests that while the fixed-length latent context captures the vast majority of information, the immense compression pressure may lead to a gradual decay of fidelity for the final portions of the source text.


\section{Conclusion and Future Work}
\label{section:conclusion}

\subsection{Conclusion}
In this work, we introduced Context Cascade Compression (C3), a novel paradigm for highly efficient long-context compression in Large Language Models. C3 proposes a more direct, pure text-to-latent pathway, implemented via a cascaded architecture of two LLMs.

Our extensive experiments on the Fox benchmark unequivocally demonstrate the superiority of this approach. C3 consistently outperforms state-of-the-art optical compression methods like Deepseek-OCR across all metrics. At compression ratios approaching 20x, where Deepseek-OCR's precision drops to approximately 60\%, C3 maintains an exceptional reconstruction precision of over 98\%. Even at extreme compression rates of nearly 40x, C3 retains a remarkable level of information fidelity. 
Furthermore, we identified a fundamental difference in the failure modes between the two paradigms. Unlike the uniform, diffuse blurring of information seen in optical compression, C3 exhibits a sequential information loss, where errors tend to concentrate at the end of the text. This behavior is compellingly analogous to the human process of memory decay, suggesting a more natural and potentially more predictable mechanism for handling information overload.

\subsection{Future Work}
The success of C3 opens up several promising avenues for future research and practical application within the LLM ecosystem.

\textbf{Enabling Ultra-long Context for LLMs:} The most immediate application of C3 is to serve as a powerful ``front-end'' compressor for existing LLMs. By compressing context that exceeds a model's native window size (e.g., millions of tokens) into a manageable number of latent tokens, C3 can unlock new capabilities in processing entire books, extensive legal documents, or large codebases for tasks like question-answering, summarization, and analysis.This will significantly reduce costs.

\textbf{Multimodal Cascaded Architecture:} Reducing costs in long-context scenarios by integrating a lightweight VLM with a LLM for processing visually-rich Documents, multi-image and video inputs.

\textbf{A Foundational Component for Next-Generation Generative Models:} C3 can function as an autoencoder architecture for encoding and decoding variable-length text. The resulting latent tokens are highly applicable to next-generation generative models, such as Diffusion Language Models and Latent Auto-Regressive Models.

\clearpage
\bibliography{neurips_2025}
\bibliographystyle{unsrtnat}

\appendix

\end{document}